\pdfoutput=1

\documentclass[11pt]{article}

\usepackage{ACL2023}

\usepackage{times}
\usepackage{latexsym}

\usepackage[LGR, T1]{fontenc}
\usepackage{alphabeta}

\usepackage[utf8]{inputenc}
\usepackage{microtype}

\usepackage[english]{babel}

\usepackage{longtable}
\usepackage{graphicx}

\usepackage{listings}
\usepackage{xcolor}
\definecolor{codegreen}{rgb}{0,0.6,0}
\definecolor{codegray}{rgb}{0.5,0.5,0.5}
\definecolor{codepurple}{rgb}{0.58,0,0.82}
\definecolor{backcolour}{rgb}{0.95,0.95,0.92}

\lstdefinestyle{mystyle}{
    backgroundcolor=\color{backcolour},   
    commentstyle=\color{codegreen},
    keywordstyle=\color{magenta},
    numberstyle=\tiny\color{codegray},
    stringstyle=\color{codepurple},
    basicstyle=\ttfamily\footnotesize,
    breakatwhitespace=false,         
    breaklines=true,                 
    captionpos=b,                    
    keepspaces=true,                 
    numbers=left,                    
    numbersep=5pt,                  
    showspaces=false,                
    showstringspaces=false,
    showtabs=false,                  
    tabsize=2
}
\title{Duncode Characters Shorter}

\author{Changshang Xue \\
  \texttt{laohur@gmail.com}
  }

\begin{document}
\maketitle

\begin{abstract}
\label{sec:abstract}
This paper investigates the employment of various encoders in text transformation, converting characters into bytes. It discusses local encoders such as ASCII and GB-2312, which encode specific characters into shorter bytes, and universal encoders like UTF-8 and UTF-16, which can encode the complete Unicode set with greater space requirements and are gaining widespread acceptance. Other encoders, including SCSU, BOCU-1, and binary encoders, however, lack self-synchronizing capabilities.

Duncode is introduced as an innovative encoding method that aims to encode the entire Unicode character set with high space efficiency, akin to local encoders. It has the potential to compress multiple characters of a string into a Duncode unit using fewer bytes. Despite offering less self-synchronizing identification information, Duncode surpasses UTF8 in terms of space efficiency. The application is available at \url{https://github.com/laohur/duncode}.

Additionally, we have developed a benchmark for evaluating character encoders across different languages. It encompasses 179 languages and can be accessed at \url{https://github.com/laohur/wiki2txt}.

\end{abstract}

\section{Introduction}
\label{sec:intro}

The process of text processing with computers begins with the encoding of characters into bytes. An ideal character encoder should possess several features such as the inclusion of all characters, compactness, and robustness (which includes lossless). However, no single character encoder embodies all these features simultaneously. Foundational encode features, such as the inclusion of all characters and robustness, are fundamental. Compactness is beneficial for conserving storage space and network bandwidth\cite{Graefe1991DataCA}.

The Unicode (ISO/IEC 10646) \cite{Unicode} coded character set, also known as the Universal Character Set (UCS) \cite{UCS}, is the largest of its kind. In the current context, the term 'including all characters' refers to the storage and transmission of Unicode characters. This is a strict requirement for global text exchange \cite{KUMARAN2003105}. Furthermore, a character encoder should ideally perform well with any language.

\textbf{Symbol Length(Bytes/Characters)} is a metric used to denote the average length of a symbol encoded into bytes within a byte sequence. This index is typically used to measure the space efficiency of encoders; however, it may vary for different characters within a single encoder.
\begin{eqnarray}
    \label{eq:SymbolLength}
    Symbol,Length = \frac {Number, of, Bytes} {Number, of, Characters}
\end{eqnarray}

Self-synchronization refers to the ability to determine whether a code unit initiates a character without referring to previous code units. This feature allows a reader to start from any point, promptly identify byte sequence boundaries, and ensure that the encoder is robust enough to exclude erroneous characters from the text. Lossy encoding is not our focus \cite{Rationale}. Some encoders specifically avoid false positives when searching for strings directly within bytes.

There are two main types of encoders used for encoding thousands of languages: local and universal encoders \cite{Nesbeitt1999EthnologueLO}. Local encoders (for example, ASCII \cite{ASCII} and ISO8859 \cite{ISO8859}) were developed earlier and typically produce shorter encoded byte lengths. However, their character sets often contain only specific symbols and are insufficient for exchange purposes. Some also lack self-synchronizing capabilities during decoding, which can lead to corruption during exchange, especially over the internet. Due to these shortcomings, the use of local encoders has declined.

Universal encoders, on the other hand, are capable of encoding the entire range of Unicode symbols. Certain encoders, such as the Unicode Transformation Format (UTF) \cite{UCS}, achieve this objective while limiting errors to a single unit, making them the most popular encoders for exchange. They typically work in conjunction with other binary compressors over the internet \cite{Pimienta2009TwelveYO}. Other universal encoders, like SCSU \cite{SCSU} and BOCU-1 \cite{BOCU}, strive to enhance encoding efficiency by introducing a tag byte as the leading byte for new blocks in a sequence. While these encoders achieve space efficiency, they do so at the expense of robustness, rendering them incapable of decoding units within a byte stream. General compressors \cite{Reference} often outperform character encoders, thanks to advanced compression algorithms \cite{Witten1987ArithmeticCF} \cite{Explained} \cite{Lelewer1987DataC}. However, these compressors lack self-synchronizing capabilities during decoding. Binary encoders, meanwhile, handle bytes after texts have already been encoded.

Most character encoders, including local encoders and UTFs, encode each character as an isolated unit. Although the character set of a text could contain millions of characters, neighboring characters often belong to the same language. These characters can be encoded using a language-specific prefix coupled with a set of letters from a small alphabet. The Unicode Point of a character can be decomposed into Alphabet ID and Letter Index within its alphabet. Subsequently, continuous symbols can be compressed by sharing a single Alphabet ID in a Duncode Unit. This shared common Alphabet ID allows Duncode to reduce the size of the encoded bytes.

In contrast to UTF-8, Duncode uses the Tail Byte (Last Byte) to segment Duncode units in the byte stream, rather than the Leading Byte (First Byte) employed by UTF-8. A Duncode unit ranges from 1-4 bytes. Given the luxury of encoding unit length into bytes, Duncode foregoes the unit length message in the segment identifier byte to maintain self-synchronizing capability. As Duncode maintains unidirectional compatibility with ASCII only, false-positive errors might occur when searching strings directly on Duncode bytes. More details on this issue are provided in Section \ref{sec:methodology} and Section \ref{sec:supplemental}.

This paper introduces a highly space-efficient text encoder – Duncoder. It represents the entire set of Unicode characters with high space efficiency and self-synchronization, albeit at the expense of occasional forbidden lookup false positives. To evaluate the compression performance of character encoders across all languages \cite{Arnold1997ACF}, we developed a tool to collect a corpus of 179 languages from Wikipedia. The results and details of these tests are presented in Section \ref{sec:experiment} and Section \ref{sec:appendix}.

\section{Related Work}
\label{sec:related}

\subsection{Local Encoder}
ASCII \cite{ASCII}, one of the most renowned encoders, encompasses a set of 128 symbols. It efficiently encodes commonly used English letters, text symbols, and computer commands into a single byte, making it particularly suitable for English language texts and the rewriting of computer commands. Western European countries often utilize the remaining 128 positions in the ASCII byte to incorporate their custom characters. Most subsequent encoders retain compatibility with ASCII. However, encoding characters from other languages often requires more bytes. For example, GB2312 \cite{GB2312} uses 2 bytes to represent thousands of commonly used Chinese characters. These local encoders usually limit their character set, leading to high space efficiency. However, conversion becomes an unavoidable obstacle in exchanges, which is far from a simple task \cite{charmod}. Unfortunately, character corruption and wide word injection are common issues in these scenarios \cite{WideCharacter}. In some extreme cases, a single erroneous byte can corrupt the entire text.

\subsection{Universal Encoder}
Character sets have become decoupled from encoders to increase universality. Unicode/ISO 10646 \cite{Unicode}, which includes over 140,000 characters and supports more than 300 languages, aims to encompass all characters globally. As a result, Unicode has become the most popular character set, and the Unicode Transformation Format (UTF) \cite{statistics} has emerged as the most widely used text encoder. As the most prevalent encoders today, UTF-8 \cite{UTF-8} and UTF-16 \cite{UTF-16} can encode all Unicode characters. UTF-16 generally has a symbol length of 2 and requires twice the space of ASCII for English. The symbol length of UTF-8 ranges from 1 to 6. UTF-8 retains ASCII in its original form and encapsulates other characters into longer bytes, making it cost-effective primarily for Latin languages.

Other universal encoders, such as SCSU \cite{SCSU} and BOCU-1 \cite{BOCU}, strive to improve encoding efficiency by inserting a tag byte as the leading byte for each new block in a sequence. However, they cannot decode units from a byte stream.

Binary compressors typically outperform character encoders, particularly those utilizing general compression algorithms \cite{Huffman1952AMF} \cite{Ziv1977AUA} \cite{Cleary1984PPM} \cite{Transform} \cite{Modeling}. However, they operate on bytes rather than characters. \cite{Making} optimizes binary compression algorithms for UTF-8 byte sequences. \cite{Fenwick1998CompressionOU} demonstrates that binary compressors effectively compress files. \cite{survey} and \cite{Matter} advocate for compression outside of text.

Some studies \cite{Manber1997ATC} \cite{Ziviani2000CompressionAK} \cite{Wan2003BrowsingAS} \cite{Takeda2001SpeedingUS} aim to accelerate text searches on compressed files.

\section{Methodology}
\label{sec:methodology}

\begin{table*}[h]
	\begin{center}
		\caption{UTF-8 Unit.  The UTF-8 unit of character "v" costs 1$\sim$6 bytes.}
		\label{tab:utf-8}
		\resizebox{\textwidth}{!}{ %

			\begin{tabular}{ccccccccccc}
				\hline
				\textbf{Bytes} & \textbf{Characters} & \textbf{Bytes/Characters} & \multicolumn{6}{c}{\textbf{Byte Sequence in Binary}}                                                        \\
				\hline
				1              & 1                   & 1                         & 0vvvvvvv                                             &          &          &          &          &          \\
				2              & 1                   & 2                         & 110vvvvv                                             & 10vvvvvv &          &          &          &          \\
				3              & 1                   & 3                         & 1110vvvv                                             & 10vvvvvv & 10vvvvvv &          &          &          \\
				4              & 1                   & 4                         & 11110vvv                                             & 10vvvvvv & 10vvvvvv & 10vvvvvv &          &          \\
				5              & 1                   & 5                         & 111110vv                                             & 10vvvvvv & 10vvvvvv & 10vvvvvv & 10vvvvvv &          \\
				6              & 1                   & 6                         & 1111110v                                             & 10vvvvvv & 10vvvvvv & 10vvvvvv & 10vvvvvv & 10vvvvvv \\
				\hline
			\end{tabular}
		}%

	\end{center}
\end{table*}

\subsection{Tail Byte for Self-Synchronizing}
In a UTF-8 byte sequence as shown in Table \ref{tab:utf-8}, only ASCII units start with the byte '0xxxxxxx' while all non-ASCII character units start with the byte "1xxxxxxx". In a multibyte unit, only the first byte resembles "11xxxxxx" and the rest resemble "10xxxxxx". The length of the "1" sequence in the first byte's head determines the number of unit bytes. The first byte of a UTF-8 unit not only segments the byte sequence but also indicates the length of this unit. However, the unit length varies only from 1 to 6. The UTF-8 unit length flag bits occupy many bits but convey a limited amount of information. Therefore, we discard this information in Duncode.

The Duncode code unit provides less synchronization information than the UTF-8 code unit. Only the \textbf{Tail Byte} (last byte of the unit) of a Duncode unit is encoded as "0xxxxxxx", with the other bytes encoded as "1xxxxxxx", as shown in Table \ref{tab:unit}. For ASCII symbols, the singular byte is also the tail byte. It is straightforward to find unit boundaries in a Duncode byte sequence. The unit length, which varies from 1 to 4, is determined dynamically by decoding the deque. Only a three-byte unit (2\textasciicircum{}21) is required to store all Unicode symbols. Each Duncode byte uses only the first bit as a flag.

We classify zones where the unit length ranges from 1 to 3 bytes. These are referred to as "ascii", "byte2", and "isolate" zones in \ref{tab:zones}, with each holding only one character per unit.

The tail byte could be an ASCII symbol or part of a longer unit, which might cause false positive errors when directly searching strings in the encoded byte sequence.

\begin{table*}[h]
	\centering
	\caption{Duncode Unit. We encode different symbols into various Duncode zones. A single character "x" can be encoded as a 1/2/3 bytes unit in zone "ascii", "byte2" and "isolate". A string of 2/3 symbols can be compressed as a 4-bytes unit for some certain languages. 3 symbols (x, y, z) of alphabet "nnn" are encoded as one unit in zone "bit8" or "bit7".
	}
	\label{tab:unit}
	\resizebox{\textwidth}{!}{ %
		\begin{tabular}{cccccccccc}
			\hline
			\textbf{ZoneId} & \textbf{ZoneName} & \multicolumn{4}{c}{\textbf{Byte Sequence in   Binary}} & \multicolumn{1}{c}{\textbf{Characters}} & \textbf{Languages} & \textbf{Bytes/Characters}                                   \\
			\hline
			0               & ascii             &                                                        &                                         &                    & 0xxxxxxx                  & x       & ascii          & 1    \\
			1               & byte2             &                                                        &                                         & 1xxxxxxx           & 0xxxxxxx                  & x       & Latin, HanZi,… & 2    \\
			2               & isolate           &                                                        & 1xxxxxxx                                & 1xxxxxxx           & 0xxxxxxx                  & x       & rare symbols   & 3    \\
			3               & bit8              & 111nnxxx                                               & 1xxxxxyy                                & 1yyyyyyz           & 0zzzzzzz                  & x, y, z & Greek…         & 1.33 \\
			4               & bit7              & 1nnnnnnn                                               & 1xxxxxxx                                & 1yyyyyyy           & 0zzzzzzz                  & x, y, z & Devanagari…    & 1.33 \\
			\hline
		\end{tabular}
	}%

\end{table*}

\subsection{Compress Multi Characters into One Duncode Unit}
Almost every self-character encoder maps a character to a single encoding unit. However, in most cases, the characters in a sentence belong to one common language. This often results in redundant information when encoding these characters as isolated units. We can deconstruct a character into two components: an \textbf{Alphabet ID} and a \textbf{Letter Index} within that alphabet. Each alphabet represents a character subset associated with a specific language, typically combining one or two Unicode blocks.

The index of a Unicode alphabet is referred to as the "Alphabet ID", while the position of a character within that alphabet is called the "Letter Index".

Multiple characters can be represented as several Letter Indexes with a shared Alphabet ID used as a prefix. Their associated alphabets typically contain no more than 128/256 letters, allowing a Letter Index to use only 7/8 bits for each alphabet. This approach enables us to compress three symbols into a 4-byte unit for these languages, reducing the symbol length to 1.33 in the "bit8" and "bit7" zones after compression.

\begin{figure*}[!h]
	\centering
	\caption{To compress the string "αβγ" into one Duncode Unit in the bit8 zone, we first decompose the string into an Alphabet ID - in this case, "Greek" - and three Letter Indexes: "0,1,2". They are then assembled into a Duncode unit within the bit8 zone. These units, ending with a Tail Byte, are easily identifiable.}
	\label{xyz}
	\includegraphics[width=1\textwidth]{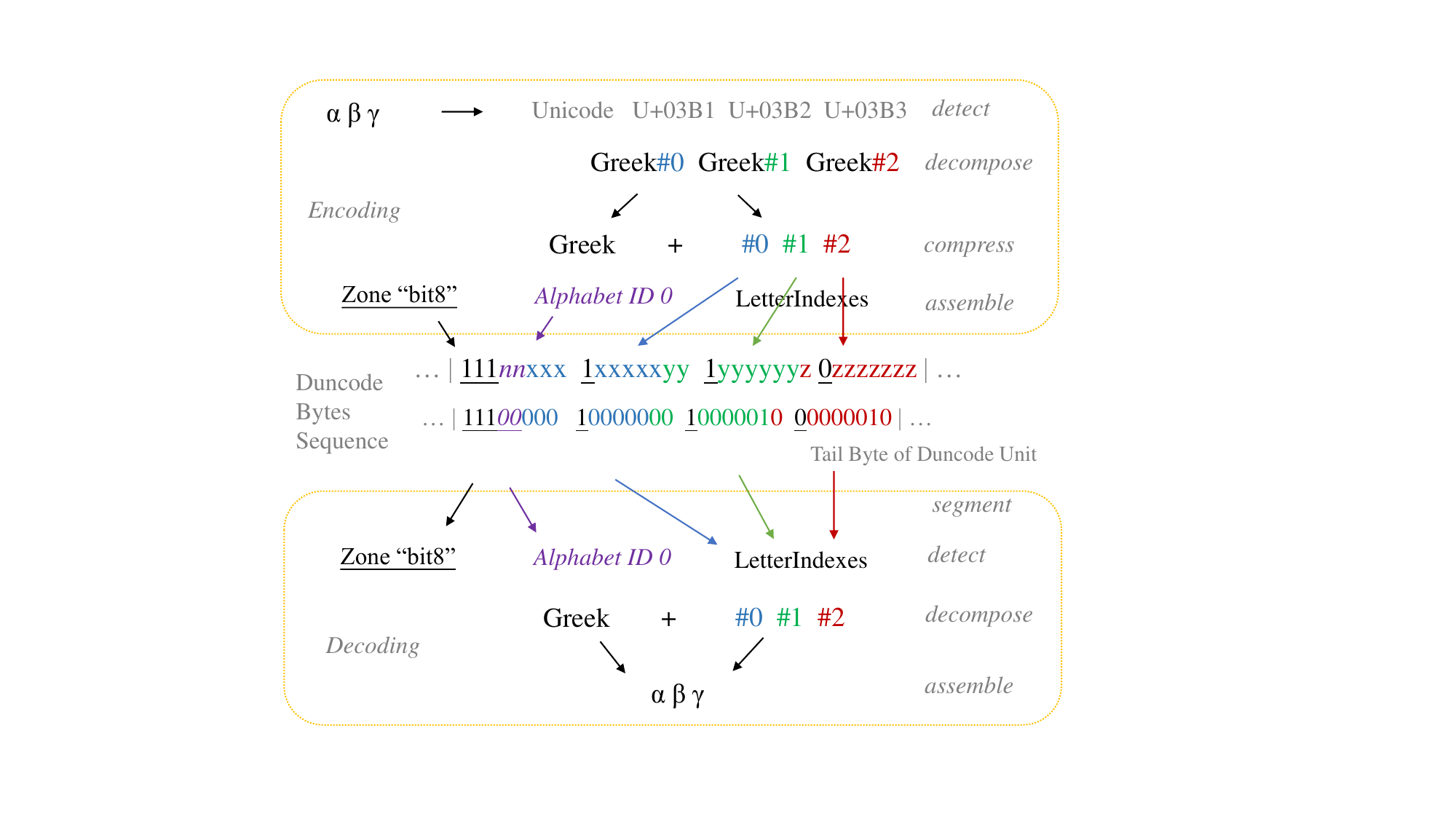}
\end{figure*}
Consider the example of encoding a Greek string “αβγ” into Duncode. Firstly, we identify that these characters are Greek, obtaining an Alphabet ID of 0 from the mapping. Subsequently, we calculate their Letter Indexes within the Greek alphabet as 0,1,2. We then assign these Alphabet ID and Letter Indexes to the Duncode bit8 zone, using Alphabet ID 1. The end result is a 4-byte Duncode unit.

\subsection{Auto Adaptive}
If a longer unit is not fully occupied by symbols, it will convert to a shorter unit. As a result, the same character may be located in different zones. For instance, the string "α" will be encoded as a two-byte unit (in the byte2 zone) instead of a four-byte unit (in the bit8 zone).

\section{Experiments and Results}
\label{sec:experiment}

\begin{table*}[h]
	\centering
	\caption{The Duncode system assigns different languages to specific zones. We encode an ASCII symbol as a one-byte unit in the ASCII zone. A frequently used character is encoded into a two-byte unit in the byte2 zone. Conversely, a rare character is encoded into a three-byte unit in the isolate zone. Around a hundred languages that utilize alphabets composed of 128 letters are encoded in the bit7 zone. Finally, four languages that use alphabets with more than 128 letters are encoded in the bit8 zone.}
	\label{tab:zones}
	\resizebox{\textwidth}{!}{ %
		\begin{tabular}{@{}ccccl@{}}
			\hline
			\textbf{Zone} & \textbf{Encode Method}                                                                                                           & \textbf{Character Sets}                                                                                   & \textbf{Capacity of Symbol Set}                         & \\
			\hline
			ascii         & =ASCII                                                                                                                           & ASCII                                                                                                     & 2\textasciicircum{}7 symbols                            & \\
			byte2         & 1-1 map                                                                                                                          & \begin{tabular}[c]{@{}c@{}}0x0080$\sim$0x07ff, hot HanZi,\\      Tibetan, Mongolian, ...\end{tabular}     & 2\textasciicircum{}14 symbols                           & \\
			isolate       & ZoneOffset + Unicode Point                                                                                                       & \begin{tabular}[c]{@{}c@{}}rare symbols, isolate   symbols,  \\      symbols of big Alphabet\end{tabular} & 2\textasciicircum{}21 symbols                           & \\
			bit8          & \begin{tabular}[c]{@{}c@{}}ZoneOffset + AlphabetOffset +   LetterIndexs,\\      then compress 3 symbols into a unit\end{tabular} & \begin{tabular}[c]{@{}c@{}}Greek and Coptic, \\      Cyrillic, Arabic, Myanmar\end{tabular}               & 4 languages of each 256-letters alphabet                & \\
			bit7          & \begin{tabular}[c]{@{}c@{}}ZoneOffset + AlphabetOffset +   LetterIndexs,\\      then compress 3 symbols into a unit\end{tabular} & common languages                                                                                          & 128 languages(Alphabets) of each   128-letters alphabet & \\
			\hline
		\end{tabular}
	}%

\end{table*}

\subsection{Zones and Alphabets }
As depicted in Table \ref{tab:zones}, we preserve ASCII characters in their original form within the ASCII zone (noted in lowercase to differentiate from ASCII).

Languages such as Latin, General Punctuation, Hiragana, Katakana, and widely-used Chinese characters are allocated to the byte2 zone, with a symbol length of 2. These are often intermixed with other languages or possess a large alphabet size that prevents compression.

Languages with alphabets containing fewer than 256 symbols can be compressed into a four-byte unit. Common languages with 128 letters or fewer are placed in the bit7 zone, while languages with more expansive alphabets (Arabic, Russian, etc., with 128-256 letters) are allocated to the bit8 zone.

Some Unicode Blocks share a single Alphabet. For instance, the Unicode Blocks "Greek and Coptic" and "Ancient Greek Numbers" share a single Duncode Alphabet. There are approximately 300 Unicode Blocks to accommodate all Unicode symbols. By uniting them, we create around 100+ Duncode Alphabets.

To determine which Alphabet a character belongs to, the Duncoder stores approximately 300 Unicode Block ranges as a list and about 2\textasciicircum{}14 characters as a map for the byte2 zone.

\subsection{Benchmark}
To evaluate compression performance across all languages, we have collected texts in 179 languages to form our corpus dataset. All corpus data for this experiment were obtained from \url{https://dumps.wikipedia.org} and extracted into texts of a maximum size of 1 MB using wiki2txt\footnote{\url{https://github.com/laohur/wiki2txt}}. Our primary comparison is between Duncoder and UTF-8, as the UTF (Unicode Transformation Format) shares many key attributes with Duncoder, and UTF-8 is the most widely used and efficient variant. The complete results are presented in Table \ref{tab:benchmark_detail}.

\begin{table*}[]
	\caption{\label{} Benchmark Duncode on Various Texts.  }
	\label{tab:benchmark}
	\resizebox{\textwidth}{!}{ %
		\begin{tabular}{@{}ccrrrccc@{}}
			\hline
			\textbf{language} & \textbf{wiki\_file\_1m} & \multicolumn{1}{c}{\textbf{n\_chars}} & \multicolumn{1}{c}{\textbf{\begin{tabular}[c]{@{}c@{}}n\_bytes\\      (utf8)\end{tabular}}} & \multicolumn{1}{c}{\textbf{\begin{tabular}[c]{@{}c@{}}n\_bytes\\      (duncode)\end{tabular}}} & \textbf{\begin{tabular}[c]{@{}c@{}}n\_bytes/n\_chars\\      (utf8)\end{tabular}} & \textbf{\begin{tabular}[c]{@{}c@{}}n\_bytes/n\_chars\\      (duncoder)\end{tabular}} & \textbf{\begin{tabular}[c]{@{}c@{}}utf8/duncode\\      (size)\end{tabular}} \\
			\hline
			English           & en.txt                  & 1,054,002                             & 1,066,474                                                                                   & 1,061,949                                                                                      & 1.01                                                                             & 1.01                                                                                 & 100.43\%                                                                    \\
			French            & fr.txt                  & 1,054,065                             & 1,096,721                                                                                   & 1,094,085                                                                                      & 1.04                                                                             & 1.04                                                                                 & 100.24\%                                                                    \\
			Arabic            & ar.txt                  & 1,103,308                             & 1,855,164                                                                                   & 1,462,890                                                                                      & 1.68                                                                             & 1.33                                                                                 & 126.82\%                                                                    \\
			Russian           & ru.txt                  & 1,049,337                             & 1,821,554                                                                                   & 1,398,275                                                                                      & 1.74                                                                             & 1.33                                                                                 & 130.27\%                                                                    \\
			Chinese           & zh.txt                  & 1,052,649                             & 2,420,113                                                                                   & 1,740,409                                                                                      & 2.30                                                                             & 1.65                                                                                 & 139.05\%                                                                    \\
			Japanese          & ja.txt                  & 1,051,113                             & 2,689,017                                                                                   & 1,872,561                                                                                      & 2.56                                                                             & 1.78                                                                                 & 143.60\%                                                                    \\
			Korean            & ko.txt                  & 1,048,759                             & 2,103,649                                                                                   & 2,087,029                                                                                      & 2.01                                                                             & 1.99                                                                                 & 100.80\%                                                                    \\
			Abkhazian         & ab.txt                  & 1,049,130                             & 1,846,144                                                                                   & 1,411,219                                                                                      & 1.76                                                                             & 1.35                                                                                 & 130.82\%                                                                    \\
			Burmese           & my.txt                  & 1,052,890                             & 2,820,647                                                                                   & 1,456,799                                                                                      & 2.68                                                                             & 1.38                                                                                 & 193.62\%                                                                    \\
			Central Khmer     & km.txt                  & 1,081,258                             & 2,890,359                                                                                   & 1,516,227                                                                                      & 2.67                                                                             & 1.40                                                                                 & 190.63\%                                                                    \\
			Tibetan           & bo.txt                  & 1,053,038                             & 3,108,029                                                                                   & 2,080,294                                                                                      & 2.95                                                                             & 1.98                                                                                 & 149.40\%                                                                    \\
			Yoruba            & yo.txt                  & 1,050,996                             & 1,230,927                                                                                   & 1,193,098                                                                                      & 1.17                                                                             & 1.14                                                                                 & 103.17\%                                                                    \\
			\hline
		\end{tabular}
	}
\end{table*}

\subsection{ Results Analysis}
The Bytes/Characters (or Symbol Length) of Duncoder in these languages align with the expected performance. In most cases, Duncoder outperforms UTF-8 across all languages. We selected several typical languages for detailed analysis, with the results provided in Table \ref{tab:benchmark}.

The performance of these encoders shows minimal variations in languages like English and French. These languages primarily consist of ASCII and Latin characters. All of these encoders encode ASCII symbols in one byte and Latin symbols in two bytes. Due to the dominance of ASCII symbols, texts in these languages are encoded at approximately one byte per character on average.

Texts in Arabic and Russian can be compressed with Duncoder, resulting in smaller sizes than UTF-8 (reducing from 2 bytes/char to 1.33 bytes/char). Due to the high prevalence of blank spaces in these texts, their average byte length per character ranges from 1 to 2 in both Duncoder and UTF-8.

Languages with extensive character sets, such as Chinese, Japanese, and Korean, are treated differently. For comparison, we have only optimized Duncoder for Hanzi (Chinese characters, CJKV Unified Ideographs), not Hangul Jamos or Syllables. The common Hanzi are stored in a map, allowing them to be encoded as two bytes. As a result, Chinese and Japanese texts undergo significant size reduction (from 3 bytes/char to 2 bytes/char). However, the size of Korean text in Duncoder remains the same (3 bytes/char) as in UTF-8.

Rare languages such as Abkhazian, Burmese, Central Khmer, Tibetan, and Yoruba all benefit from Duncoder, with the exception of Yoruba. This is because Yoruba primarily uses ASCII and Latin symbols. Whether a language benefits from this approach depends on its character set distribution. Abkhazian primarily uses Cyrillic, while Burmese, Central Khmer, and Tibetan use their unique character sets. As a result, these languages reap significant benefits from Duncoder (with byte/char values shifting from [2,3] to [1.33,2]).

\section{Conclusion}
\label{sec:conclusion}

In this paper, we introduced Duncode to encode various texts in high space efficiency. It is universal, robust, customizable and keep the unidirectional compatibility with ASCII. With the cost of lookup false positive, Duncode benefits a great space advantage than UTF-8. It is particular suitable for storing diverse texts. In addition, we released a corpus including 179 languages and a tool to collect them. And we built a benchmark for character encoders on all languages.

\bibliography{bibliography/duncode}
\bibliographystyle{acl_natbib}

\appendix
\newpage
\section{Appendices}
\label{sec:appendix}

\subsection{Encoding Steps}

\paragraph{Detect Alphabet ID}
The Alphabet may encompass one or two Unicode Blocks. We use a lookup table to identify to which Alphabet a given character belongs.

\paragraph{Decompose}
Next, we decompose a character from a Unicode Point into an Alphabet ID and an Index within the Alphabet.
The Unicode Point is equal to the sum of the Alphabet ID and the Letter Index.

\paragraph{Compress}
In an uncompressed Duncode unit sequence, if the symbol in the last unit and the current symbol belong to the same alphabet and the last unit is not full, the current symbol will be inserted into the last unit, and the current Duncode unit will be discarded.
As depicted in Figure \ref{xyz}, we insert 3 symbols into one Duncode unit.

\paragraph{Assemble}
For each compressed Duncode unit, we assemble the Alphabet ID and Indexes into different Duncode Zones based on the Alphabet ID.

\subsection{Decoding Steps}

\paragraph{Segment}
For a Duncode byte sequence, Duncode units can be split by the Tail Byte (0xxxxxxx).

\paragraph{Detect Zone, Alphabet ID, and Character Indexes}
For each compressed Duncode unit, we first detect the zone using flag bits. The Alphabet ID and Character Indexes are identified subsequently.

\paragraph{Decompress}
We decompress a Duncode unit containing multiple Letter Indexes into several Duncode units, ensuring that each unit contains only one symbol.

\paragraph{Generate Character}
For each decompressed Duncode unit, we can assemble the Alphabet ID and Letter Index into a character in the appropriate zone.

\onecolumn
\begin{longtable}{@{}crrrll@{}}
    \caption{\label{} Benchmark Details on corpus. "hz.txt" and "kr.txt" are empty after extracted. }
    \label{tab:benchmark_detail}                                                                                                                                     \\

    \multicolumn{1}{l}{\textbf{wiki\_file\_1m}} & \textbf{n\_chars} & \textbf{bytes\_utf8} & \textbf{bytes\_duncode} & \textbf{bytes/utf8} & \textbf{bytes/duncoder} \\
    aa.txt                                      & 135               & 148                  & 147                     & 1.096               & 1.089                   \\
    ab.txt                                      & 1,049,130         & 1,846,144            & 1,411,219               & 1.760               & 1.345                   \\
    af.txt                                      & 1,064,351         & 1,075,107            & 1,071,541               & 1.010               & 1.007                   \\
    ak.txt                                      & 769,002           & 814,811              & 812,865                 & 1.060               & 1.057                   \\
    am.txt                                      & 1,050,665         & 2,437,382            & 2,431,631               & 2.320               & 2.314                   \\
    an.txt                                      & 1,051,327         & 1,068,238            & 1,066,376               & 1.016               & 1.014                   \\
    ar.txt                                      & 1,103,308         & 1,855,164            & 1,462,890               & 1.681               & 1.326                   \\
    as.txt                                      & 1,052,400         & 2,262,353            & 1,376,996               & 2.150               & 1.308                   \\
    av.txt                                      & 1,048,809         & 1,774,760            & 1,368,398               & 1.692               & 1.305                   \\
    ay.txt                                      & 1,049,302         & 1,069,827            & 1,066,001               & 1.020               & 1.016                   \\
    az.txt                                      & 1,048,850         & 1,211,125            & 1,201,764               & 1.155               & 1.146                   \\
    ba.txt                                      & 1,048,631         & 1,850,444            & 1,403,582               & 1.765               & 1.338                   \\
    be.txt                                      & 1,048,579         & 1,825,171            & 1,402,696               & 1.741               & 1.338                   \\
    bg.txt                                      & 1,049,268         & 1,777,473            & 1,400,699               & 1.694               & 1.335                   \\
    bh.txt                                      & 1,048,911         & 2,190,884            & 1,380,707               & 2.089               & 1.316                   \\
    bi.txt                                      & 218,091           & 221,344              & 219,201                 & 1.015               & 1.005                   \\
    bm.txt                                      & 357,032           & 378,550              & 376,310                 & 1.060               & 1.054                   \\
    bn.txt                                      & 1,060,806         & 2,472,105            & 1,450,265               & 2.330               & 1.367                   \\
    bo.txt                                      & 1,053,038         & 3,108,029            & 2,080,294               & 2.951               & 1.976                   \\
    br.txt                                      & 1,055,652         & 1,079,657            & 1,075,782               & 1.023               & 1.019                   \\
    bs.txt                                      & 1,049,786         & 1,072,485            & 1,070,267               & 1.022               & 1.020                   \\
    ca.txt                                      & 1,073,230         & 1,100,117            & 1,098,909               & 1.025               & 1.024                   \\
    ce.txt                                      & 1,048,602         & 1,837,755            & 1,406,538               & 1.753               & 1.341                   \\
    ch.txt                                      & 129,361           & 134,508              & 132,508                 & 1.040               & 1.024                   \\
    co.txt                                      & 1,060,056         & 1,090,727            & 1,088,470               & 1.029               & 1.027                   \\
    cr.txt                                      & 17,333            & 28,113               & 27,833                  & 1.622               & 1.606                   \\
    cs.txt                                      & 1,055,737         & 1,159,181            & 1,153,879               & 1.098               & 1.093                   \\
    cu.txt                                      & 462,741           & 825,969              & 652,117                 & 1.785               & 1.409                   \\
    cv.txt                                      & 1,049,285         & 1,844,550            & 1,466,533               & 1.758               & 1.398                   \\
    cy.txt                                      & 1,050,639         & 1,060,864            & 1,058,520               & 1.010               & 1.008                   \\
    da.txt                                      & 1,050,196         & 1,077,241            & 1,075,205               & 1.026               & 1.024                   \\
    de.txt                                      & 1,050,528         & 1,072,438            & 1,068,837               & 1.021               & 1.017                   \\
    dv.txt                                      & 1,050,848         & 1,921,186            & 1,416,724               & 1.828               & 1.348                   \\
    dz.txt                                      & 310,724           & 858,221              & 584,403                 & 2.762               & 1.881                   \\
    ee.txt                                      & 280,228           & 297,292              & 295,492                 & 1.061               & 1.054                   \\
    el.txt                                      & 1,049,405         & 1,770,860            & 1,369,767               & 1.687               & 1.305                   \\
    en.txt                                      & 1,054,002         & 1,066,474            & 1,061,949               & 1.012               & 1.008                   \\
    eo.txt                                      & 1,048,873         & 1,078,033            & 1,073,017               & 1.028               & 1.023                   \\
    es.txt                                      & 1,049,707         & 1,075,445            & 1,072,754               & 1.025               & 1.022                   \\
    et.txt                                      & 1,049,247         & 1,085,530            & 1,079,530               & 1.035               & 1.029                   \\
    eu.txt                                      & 1,048,842         & 1,060,860            & 1,057,453               & 1.011               & 1.008                   \\
    fa.txt                                      & 1,058,698         & 1,816,881            & 1,457,016               & 1.716               & 1.376                   \\
    ff.txt                                      & 503,152           & 617,832              & 538,409                 & 1.228               & 1.070                   \\
    fi.txt                                      & 1,050,348         & 1,085,591            & 1,082,824               & 1.034               & 1.031                   \\
    fj.txt                                      & 497,382           & 500,275              & 498,381                 & 1.006               & 1.002                   \\
    fo.txt                                      & 1,049,176         & 1,122,285            & 1,120,126               & 1.070               & 1.068                   \\
    fr.txt                                      & 1,054,065         & 1,096,721            & 1,094,085               & 1.040               & 1.038                   \\
    fy.txt                                      & 1,065,075         & 1,080,793            & 1,079,544               & 1.015               & 1.014                   \\
    ga.txt                                      & 1,048,950         & 1,106,459            & 1,103,747               & 1.055               & 1.052                   \\
    gd.txt                                      & 1,048,838         & 1,086,757            & 1,078,849               & 1.036               & 1.029                   \\
    gl.txt                                      & 1,050,492         & 1,077,723            & 1,076,035               & 1.026               & 1.024                   \\
    gn.txt                                      & 1,051,576         & 1,131,966            & 1,118,462               & 1.076               & 1.064                   \\
    gu.txt                                      & 1,065,827         & 2,499,704            & 1,448,002               & 2.345               & 1.359                   \\
    gv.txt                                      & 1,049,881         & 1,060,374            & 1,057,448               & 1.010               & 1.007                   \\
    ha.txt                                      & 1,049,904         & 1,059,909            & 1,057,872               & 1.010               & 1.008                   \\
    he.txt                                      & 1,058,309         & 1,833,281            & 1,434,241               & 1.732               & 1.355                   \\
    hi.txt                                      & 1,049,758         & 2,608,190            & 1,487,044               & 2.485               & 1.417                   \\
    ho.txt                                      & 1,218             & 1,227                & 1,220                   & 1.007               & 1.002                   \\
    hr.txt                                      & 1,051,227         & 1,079,772            & 1,077,857               & 1.027               & 1.025                   \\
    ht.txt                                      & 1,050,145         & 1,070,733            & 1,069,807               & 1.020               & 1.019                   \\
    hu.txt                                      & 1,051,729         & 1,148,520            & 1,144,229               & 1.092               & 1.088                   \\
    hy.txt                                      & 1,052,342         & 1,811,174            & 1,385,451               & 1.721               & 1.317                   \\
    hz.txt                                      & 0                 & 0                    & 0                       & \-                  & \-                      \\
    ia.txt                                      & 1,049,441         & 1,057,971            & 1,053,944               & 1.008               & 1.004                   \\
    id.txt                                      & 1,051,882         & 1,063,624            & 1,059,494               & 1.011               & 1.007                   \\
    ie.txt                                      & 1,048,912         & 1,146,682            & 1,110,250               & 1.093               & 1.058                   \\
    ig.txt                                      & 1,052,913         & 1,166,756            & 1,113,920               & 1.108               & 1.058                   \\
    ii.txt                                      & 261               & 478                  & 415                     & 1.831               & 1.590                   \\
    ik.txt                                      & 83,095            & 87,540               & 86,225                  & 1.053               & 1.038                   \\
    io.txt                                      & 1,049,261         & 1,055,452            & 1,053,891               & 1.006               & 1.004                   \\
    is.txt                                      & 1,049,702         & 1,149,928            & 1,146,471               & 1.095               & 1.092                   \\
    it.txt                                      & 1,064,336         & 1,074,815            & 1,072,780               & 1.010               & 1.008                   \\
    iu.txt                                      & 102,983           & 216,868              & 215,287                 & 2.106               & 2.091                   \\
    ja.txt                                      & 1,051,113         & 2,689,017            & 1,872,561               & 2.558               & 1.782                   \\
    jv.txt                                      & 1,048,611         & 1,071,004            & 1,066,499               & 1.021               & 1.017                   \\
    ka.txt                                      & 1,054,859         & 2,534,120            & 1,412,512               & 2.402               & 1.339                   \\
    kg.txt                                      & 235,777           & 245,142              & 240,662                 & 1.040               & 1.021                   \\
    ki.txt                                      & 289,476           & 311,073              & 309,083                 & 1.075               & 1.068                   \\
    kj.txt                                      & 838               & 849                  & 840                     & 1.013               & 1.002                   \\
    kk.txt                                      & 1,049,332         & 1,884,244            & 1,411,157               & 1.796               & 1.345                   \\
    kl.txt                                      & 402,490           & 404,972              & 403,637                 & 1.006               & 1.003                   \\
    km.txt                                      & 1,081,258         & 2,890,359            & 1,516,227               & 2.673               & 1.402                   \\
    kn.txt                                      & 1,051,506         & 2,709,242            & 1,448,139               & 2.577               & 1.377                   \\
    ko.txt                                      & 1,048,759         & 2,103,649            & 2,087,029               & 2.006               & 1.990                   \\
    kr.txt                                      & 0                 & 0                    & 0                       & \-                  & \-                      \\
    ks.txt                                      & 358,381           & 619,065              & 461,352                 & 1.727               & 1.287                   \\
    ku.txt                                      & 1,050,172         & 1,149,542            & 1,145,047               & 1.095               & 1.090                   \\
    kv.txt                                      & 1,049,872         & 1,744,416            & 1,404,876               & 1.662               & 1.338                   \\
    kw.txt                                      & 1,059,998         & 1,068,603            & 1,064,158               & 1.008               & 1.004                   \\
    ky.txt                                      & 1,072,845         & 1,914,993            & 1,441,986               & 1.785               & 1.344                   \\
    la.txt                                      & 1,057,597         & 1,069,208            & 1,065,448               & 1.011               & 1.007                   \\
    lb.txt                                      & 1,051,739         & 1,076,859            & 1,074,987               & 1.024               & 1.022                   \\
    lg.txt                                      & 1,049,053         & 1,071,739            & 1,060,447               & 1.022               & 1.011                   \\
    li.txt                                      & 1,048,897         & 1,066,137            & 1,065,019               & 1.016               & 1.015                   \\
    ln.txt                                      & 1,049,167         & 1,115,033            & 1,110,845               & 1.063               & 1.059                   \\
    lo.txt                                      & 1,051,413         & 2,656,032            & 1,383,759               & 2.526               & 1.316                   \\
    lt.txt                                      & 1,054,230         & 1,132,785            & 1,121,031               & 1.075               & 1.063                   \\
    lv.txt                                      & 1,050,215         & 1,150,328            & 1,139,656               & 1.095               & 1.085                   \\
    mg.txt                                      & 1,049,224         & 1,064,549            & 1,061,628               & 1.015               & 1.012                   \\
    mh.txt                                      & 3,148             & 3,270                & 3,261                   & 1.039               & 1.036                   \\
    mi.txt                                      & 1,048,616         & 1,090,831            & 1,088,423               & 1.040               & 1.038                   \\
    mk.txt                                      & 1,050,407         & 1,807,308            & 1,412,958               & 1.721               & 1.345                   \\
    ml.txt                                      & 1,049,136         & 2,564,675            & 1,389,893               & 2.445               & 1.325                   \\
    mn.txt                                      & 1,050,075         & 1,828,576            & 1,395,636               & 1.741               & 1.329                   \\
    mr.txt                                      & 1,051,589         & 2,560,750            & 1,441,600               & 2.435               & 1.371                   \\
    ms.txt                                      & 1,049,273         & 1,056,111            & 1,053,374               & 1.007               & 1.004                   \\
    mt.txt                                      & 1,052,893         & 1,095,095            & 1,090,599               & 1.040               & 1.036                   \\
    my.txt                                      & 1,052,890         & 2,820,647            & 1,456,799               & 2.679               & 1.384                   \\
    na.txt                                      & 224,739           & 234,089              & 229,751                 & 1.042               & 1.022                   \\
    ne.txt                                      & 1,052,018         & 2,634,100            & 1,473,330               & 2.504               & 1.400                   \\
    ng.txt                                      & 21,339            & 21,378               & 21,352                  & 1.002               & 1.001                   \\
    nl.txt                                      & 1,189,301         & 1,194,604            & 1,192,975               & 1.004               & 1.003                   \\
    nn.txt                                      & 1,056,326         & 1,087,947            & 1,083,779               & 1.030               & 1.026                   \\
    no.txt                                      & 1,052,080         & 1,079,185            & 1,075,334               & 1.026               & 1.022                   \\
    nv.txt                                      & 1,048,735         & 1,345,272            & 1,340,291               & 1.283               & 1.278                   \\
    ny.txt                                      & 1,049,312         & 1,055,213            & 1,053,021               & 1.006               & 1.004                   \\
    oc.txt                                      & 1,049,379         & 1,080,530            & 1,078,600               & 1.030               & 1.028                   \\
    om.txt                                      & 1,052,694         & 1,061,522            & 1,057,797               & 1.008               & 1.005                   \\
    or.txt                                      & 1,049,082         & 2,287,591            & 1,380,361               & 2.181               & 1.316                   \\
    os.txt                                      & 1,048,871         & 1,860,170            & 1,485,113               & 1.773               & 1.416                   \\
    pa.txt                                      & 1,050,021         & 2,410,900            & 1,462,044               & 2.296               & 1.392                   \\
    pi.txt                                      & 473,902           & 777,614              & 552,513                 & 1.641               & 1.166                   \\
    pl.txt                                      & 1,088,573         & 1,152,076            & 1,143,677               & 1.058               & 1.051                   \\
    ps.txt                                      & 1,057,349         & 1,815,710            & 1,471,991               & 1.717               & 1.392                   \\
    pt.txt                                      & 1,054,300         & 1,085,823            & 1,084,075               & 1.030               & 1.028                   \\
    qu.txt                                      & 1,049,115         & 1,071,419            & 1,068,046               & 1.021               & 1.018                   \\
    rm.txt                                      & 1,075,028         & 1,103,281            & 1,094,257               & 1.026               & 1.018                   \\
    rn.txt                                      & 359,070           & 365,332              & 362,715                 & 1.017               & 1.010                   \\
    ro.txt                                      & 1,048,802         & 1,099,145            & 1,094,233               & 1.048               & 1.043                   \\
    ru.txt                                      & 1,049,337         & 1,821,554            & 1,398,275               & 1.736               & 1.333                   \\
    rw.txt                                      & 1,048,887         & 1,062,604            & 1,056,094               & 1.013               & 1.007                   \\
    sa.txt                                      & 1,052,407         & 2,655,317            & 1,446,092               & 2.523               & 1.374                   \\
    sc.txt                                      & 1,048,935         & 1,074,319            & 1,068,526               & 1.024               & 1.019                   \\
    sd.txt                                      & 1,048,901         & 1,761,417            & 1,421,304               & 1.679               & 1.355                   \\
    se.txt                                      & 1,048,913         & 1,126,991            & 1,109,431               & 1.074               & 1.058                   \\
    sg.txt                                      & 157,336           & 169,263              & 168,661                 & 1.076               & 1.072                   \\
    sh.txt                                      & 1,050,057         & 1,072,112            & 1,068,282               & 1.021               & 1.017                   \\
    si.txt                                      & 1,049,987         & 2,429,987            & 1,425,067               & 2.314               & 1.357                   \\
    sk.txt                                      & 1,049,200         & 1,134,432            & 1,131,023               & 1.081               & 1.078                   \\
    sl.txt                                      & 1,052,257         & 1,082,072            & 1,078,131               & 1.028               & 1.025                   \\
    sm.txt                                      & 762,554           & 774,167              & 770,192                 & 1.015               & 1.010                   \\
    sn.txt                                      & 1,049,047         & 1,052,745            & 1,050,726               & 1.004               & 1.002                   \\
    so.txt                                      & 1,049,058         & 1,055,824            & 1,053,031               & 1.006               & 1.004                   \\
    sq.txt                                      & 1,048,696         & 1,114,188            & 1,111,205               & 1.062               & 1.060                   \\
    sr.txt                                      & 1,050,605         & 1,717,480            & 1,364,645               & 1.635               & 1.299                   \\
    ss.txt                                      & 546,788           & 554,288              & 551,570                 & 1.014               & 1.009                   \\
    st.txt                                      & 817,347           & 822,214              & 820,598                 & 1.006               & 1.004                   \\
    su.txt                                      & 1,048,681         & 1,085,020            & 1,082,192               & 1.035               & 1.032                   \\
    sv.txt                                      & 1,052,621         & 1,104,884            & 1,097,727               & 1.050               & 1.043                   \\
    sw.txt                                      & 1,049,738         & 1,055,492            & 1,052,862               & 1.005               & 1.003                   \\
    ta.txt                                      & 1,048,999         & 2,460,455            & 1,374,203               & 2.346               & 1.310                   \\
    te.txt                                      & 1,066,784         & 2,682,043            & 1,459,958               & 2.514               & 1.369                   \\
    tg.txt                                      & 1,048,631         & 1,773,182            & 1,382,015               & 1.691               & 1.318                   \\
    th.txt                                      & 1,049,536         & 2,747,511            & 1,385,799               & 2.618               & 1.320                   \\
    ti.txt                                      & 158,935           & 377,391              & 376,330                 & 2.374               & 2.368                   \\
    tk.txt                                      & 1,048,962         & 1,159,714            & 1,152,289               & 1.106               & 1.099                   \\
    tl.txt                                      & 1,049,152         & 1,060,988            & 1,055,790               & 1.011               & 1.006                   \\
    tn.txt                                      & 1,048,609         & 1,052,937            & 1,051,558               & 1.004               & 1.003                   \\
    to.txt                                      & 915,162           & 962,213              & 960,159                 & 1.051               & 1.049                   \\
    tr.txt                                      & 1,049,314         & 1,136,726            & 1,134,037               & 1.083               & 1.081                   \\
    ts.txt                                      & 975,843           & 983,577              & 980,917                 & 1.008               & 1.005                   \\
    tt.txt                                      & 1,048,655         & 1,686,413            & 1,336,501               & 1.608               & 1.274                   \\
    tw.txt                                      & 1,048,945         & 1,112,019            & 1,109,909               & 1.060               & 1.058                   \\
    ty.txt                                      & 149,036           & 166,322              & 160,290                 & 1.116               & 1.076                   \\
    ug.txt                                      & 1,051,124         & 1,950,004            & 1,477,330               & 1.855               & 1.405                   \\
    uk.txt                                      & 1,048,828         & 1,800,824            & 1,391,203               & 1.717               & 1.326                   \\
    ur.txt                                      & 1,051,569         & 1,784,925            & 1,444,569               & 1.697               & 1.374                   \\
    uz.txt                                      & 1,049,363         & 1,088,618            & 1,075,666               & 1.037               & 1.025                   \\
    ve.txt                                      & 242,125           & 245,211              & 243,424                 & 1.013               & 1.005                   \\
    vi.txt                                      & 1,173,008         & 1,501,596            & 1,388,612               & 1.280               & 1.184                   \\
    vo.txt                                      & 1,050,168         & 1,145,071            & 1,132,800               & 1.090               & 1.079                   \\
    wa.txt                                      & 1,050,692         & 1,095,828            & 1,092,625               & 1.043               & 1.040                   \\
    wo.txt                                      & 1,048,791         & 1,091,114            & 1,087,101               & 1.040               & 1.037                   \\
    xh.txt                                      & 1,087,868         & 1,092,314            & 1,090,715               & 1.004               & 1.003                   \\
    yi.txt                                      & 1,051,166         & 1,849,231            & 1,416,118               & 1.759               & 1.347                   \\
    yo.txt                                      & 1,050,996         & 1,230,927            & 1,193,098               & 1.171               & 1.135                   \\
    za.txt                                      & 505,418           & 606,031              & 555,463                 & 1.199               & 1.099                   \\
    zh.txt                                      & 1,052,649         & 2,420,113            & 1,740,409               & 2.299               & 1.653                   \\
    zu.txt                                      & 1,050,600         & 1,059,590            & 1,054,789               & 1.009               & 1.004                   \\
\end{longtable}

\newpage
\section{Supplemental Material}
\label{sec:supplemental}

\subsection{Duncode Algorithm}

\paragraph{Duncode Block DataStructure}
A Duncode Block bears a resemblance to a Unicode Block. Typically, one Duncode Block corresponds to one Unicode Block. However, at times, two smaller Unicode Blocks are consolidated to constitute one Duncode Block. In such scenarios, the initial block is termed the Mother Block, while the subsequent ones are identified as Children Blocks.

\begin{lstlisting}[  
    language=Go, 
    breaklines=true, 
   ]  
type Block struct {
    BlockId  int  // Index of the Duncode Block
    Began    int  // Unicode Point of the first symbol in this Duncode Block
    End      int  // Unicode Point of the last symbol in this Duncode Block
    Size     int  // Symbol Capacity of this Duncode Block
    English  string // English name of the Block
    Chinese  string  // Chinese name of the Block
    Mother   string  // Name of the Mother Duncode Block for a Child Duncode Block
    MotherId int  //  ID of the Mother Duncode Block for a Child Duncode Block
    Offset   int  // For a Child Duncode Block, Block Offset = Beginning of Mother Block + Block Size
    Child    []string   // Names of Child Duncode Blocks
    ZoneName string  // Name of the Zone
    ZoneId   int  // ID of the Zone
    Zone2Id  int  // ID of the 'bit7' Zone
    Zone3Id  int  // ID of the 'bit8' Zone
}
\end{lstlisting}

\paragraph{Duncode Unit DataStructure}
The Duncode struct represents a Duncode Unit, which can store 1 to 3 symbols from the same Duncode Block after compression.
\begin{lstlisting}[language=Go]
type Duncode struct {
    CodePoint int  // ID of the Duncode Zone
    ZoneId    int  // ID of the Duncode Zone
    BlockId   int  // ID of the Duncode Block
    MotherId  int  // ID of the Mother Duncode Block
    Index     int  // Index of the Duncode Unit = Unicode CodePoint - Beginning of the Duncode Block
    Symbols   []int  // Array storing 1 to 3 Duncode Indexes in compressed Duncode Units
}
\end{lstlisting}



\end{document}